\pdfoutput=1
\documentclass[10pt,twocolumn,letterpaper]{article}
\usepackage[pagenumbers]{cvpr}
\usepackage{graphicx}
\usepackage{amsmath,amssymb}
\usepackage{booktabs}
\usepackage{xcolor}
\newcommand{\method}{MoReWorld}

\definecolor{cvprblue}{rgb}{0.21,0.49,0.74}
\usepackage[pagebackref,breaklinks,colorlinks,allcolors=cvprblue]{hyperref}
\title{Learning Interaction Dynamics via Motion-Region Guidance \\
for Robotic Video World Models}
\author{
  YongKang He$^{1}$, Xiaofen Xing$^{1,*}$, XiangMin Xu$^{2,*}$\\
  $^{1}$South China University of Technology \quad
  $^{2}$Foshan University \quad
  $^{*}$Corresponding author \\
  {\tt\small heeeyk99@gmail.com, xfxing@scut.edu.cn, xmxu@scut.edu.cn}
}

\begin{document}
\maketitle
\begin{abstract}
Action-conditioned video world models imagine future observations under candidate robot actions, providing a promising visual tool for robot policy evaluation and interaction simulation.
However, existing video world models often optimize global reconstruction errors instead of emphasizing interaction-relevant regions. As a result, they may preserve overall scene appearance but overlook local interaction details, leading to physically implausible outcomes such as hallucinated contact effects and inconsistent robot–object interactions.
We introduce MoReWorld, a unified motion-region-guided framework that uses a shared motion region, the part of the environment that moves in response to the robot's actions, to focus dynamics learning and video generation on interaction-relevant regions. The same motion region guides local motion prediction, region-aware supervision, and motion-gated future-feature conditioning. At inference time, the motion region is predicted as a motion probability map from visual history and robot actions, without requiring future observations or ground-truth motion labels.
On WorldSimProbe, MoReWorld achieves the highest overall simulator-faithfulness score, with the strongest performance on local consistency, global consistency, and interaction dynamics. These results indicate that motion-region guidance helps the model preserve scene-level consistency while more faithfully predicting action-conditioned interaction dynamics.

\end{abstract}

\section{Introduction}
\label{sec:introduction}

\begin{figure}[t]
\centering
\includegraphics[width=\linewidth]{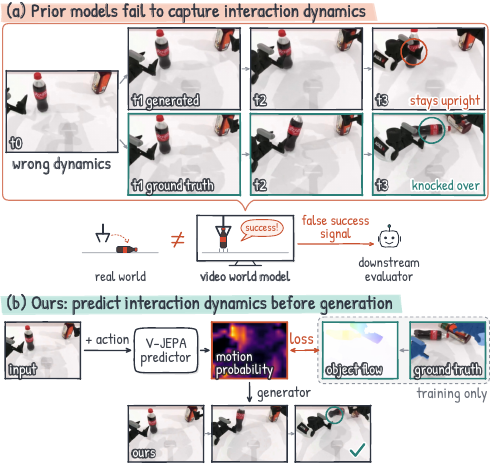}
\caption{ Accurate interaction-dynamics prediction is critical for video world models. (a) Prior models can produce incorrect object responses, yielding false success signals that mislead downstream evaluation. (b) MoReWorld predicts interaction dynamics before generation via a motion-probability intermediate, producing object motion and a final video consistent with the action.}
\label{fig:motivation}
\end{figure}

Action-conditioned video world models offer a promising foundation for embodied simulation: given visual observations and robot actions, they can synthesize future interactions for visual planning, policy evaluation, and policy learning~\cite{guo2025ctrlworld,gigaworld2026,jiang2026wovr,shang2026worldarena2}. Their visual priors make diverse scenes accessible to learned simulation, but visual plausibility alone does not guarantee faithful action consequences. As illustrated in Fig.~\ref{fig:motivation}(a), prior models may fail to capture correct physical dynamics, e.g., the manipulated object remains upright when it should be knocked over. Such failures indicate that the model has reproduced scene appearance without reliably modeling the interaction dynamics induced by the action.

When generated videos serve as simulated experience in closed-loop policy learning, an erroneously generated observation can change the next action and thus propagate the error throughout the rollout, ultimately affecting the reward. Such errors can undermine their reliability as simulated experience. As shown in the lower part of Fig.~\ref{fig:motivation}(a), an apparently successful trajectory may be merely a generation artifact, yielding a false success signal that misleads downstream evaluation. WoVR provides evidence for this concern by showing how hallucinated outcomes and accumulated prediction errors corrupt imagined learning signals~\cite{jiang2026wovr}. Relatedly, policy-oriented benchmarks emphasize faithful rollouts and functional utility beyond conventional video-quality metrics~\cite{gigaworld2026,shang2026worldarena2,co2026worldsimprobe}. These considerations point to a central requirement for action-conditioned video world models: \emph{action-consequence fidelity}, i.e., the generated robot motion and the resulting environmental changes must remain consistent with the supplied actions, even for unsuccessful interactions.

Existing approaches have tackled various aspects of this requirement. Visual, kinematic, and contact-based action interfaces provide generators with explicit control over robot motion~\cite{alzayer2026masked,kim2026robotfactored,xu2026kinema4d,azirar2026contactflow}. Supervision methods based on geometry, trajectories, relations, and action-sensitivity further inject structural priors to emphasize interaction-relevant elements, such as re-weighting spatial features via causal effects~\cite{dreamx2026,zhang2026physisforcing,fang2026caer}. Explicit generative coupling further connects predicted dynamics to visual synthesis: FlowDreamer conditions image generation on scene flow, while PHANTOM jointly evolves visual and latent physical representations via cross-attention~\cite{guo2025flowdreamer,shen2026phantom}. Despite these efforts, existing methods typically treat interaction signals either as implicit feature weights or direct external conditions. They do not explicitly predict the spatial regions where interactions will occur and subsequently use this internally predicted prior to actively guide the generation process. This raises a more specific question: how can an action-conditioned video model explicitly identify interaction-relevant regions and leverage their predicted motion to preserve both local interaction changes and global scene consistency?

To address this question, we propose \textbf{MoReWorld}, a motion-region-guided action-conditioned video world model. As depicted in Fig.~\ref{fig:motivation}(b), MoReWorld explicitly predicts motion before generation. It organizes learning and generation around a single motion region, realized through two complementary mechanisms. During inference, the model explicitly predicts a motion probability map from the observed history, actions, and robot states to guide the generation process. During training, it constructs fine-grained motion-map labels—by excluding robot-covered regions and separating moving from static environments—to re-weight the supervision and focus learning on interaction-relevant dynamics. This shared motion region plays a triple role: it supervises local motion prediction, dynamically balances the loss between moving and static regions, and explicitly guides the denoising process to inject future features only where necessary. This explicit \emph{Motion-Region Guidance} enables MoReWorld to accurately learn Interaction Dynamics—emphasizing local interaction-relevant changes while preserving the static scene evidence needed for globally coherent video synthesis.

We evaluate MoReWorld on WorldSimProbe, a benchmark that measures simulator faithfulness through local consistency, global consistency, and interaction dynamics. MoReWorld achieves the strongest overall simulator-faithfulness score and significantly improves interaction-oriented criteria, demonstrating the effectiveness of spatially organized motion guidance.

Our contributions are threefold:
\begin{itemize}
\item We introduce MoReWorld, a motion-region-guided action-conditioned video world model that organizes interaction dynamics learning and future video generation around spatially localized environmental changes.
\item We develop a unified motion-region pipeline that combines action-conditioned motion probability prediction, optical-flow and robot-mask-based motion-map labels, and motion-gated future-feature conditioning in a shared video generation backbone.
\item We propose region-aware supervision that balances moving and static environment regions while excluding robot-covered areas, and demonstrate improved simulator faithfulness, local consistency, global consistency, and interaction dynamics on WorldSimProbe.
\end{itemize}
\section{Related Work}
\label{sec:related_work}

\subsection{Video Diffusion Models as World Models}
Pretrained video diffusion models provide visual and temporal priors for predicting how a scene evolves~\cite{wan2025,wan22}.
Extending these generators toward interactive world models requires controllability and persistent context beyond short clips.
Recent approaches explore external scene memory and long-horizon supervision~\cite{yin2026evoke}, geometry-aware action and memory interfaces~\cite{yan2026egogenesis}, and streaming generation controlled by three-dimensional point tracks~\cite{li2026streamctrl}.
In robotics, action-conditioned models further adapt video generation to repeated policy--environment interaction, with multi-view control and memory supporting imagined rollouts~\cite{guo2025ctrlworld,dreamx2026}.
Their usefulness must be assessed beyond visual quality: Ctrl-World and GigaWorld-1 study policy evaluation~\cite{guo2025ctrlworld,gigaworld2026}, while WoVR and WorldArena~2.0 examine learned simulators for policy improvement~\cite{jiang2026wovr,shang2026worldarena2}.

Video-based experience also supports robot learning by extracting executable object-motion goals from generated videos~\cite{dharmarajan2025dream2flow}.
Our focus is the forward prediction problem underlying embodied simulation: given prescribed robot actions, generate the resulting robot and object evolution.
We investigate how internally predicted motion should influence RGB denoising, rather than introducing a new policy-learning pipeline or persistent-memory architecture.

\subsection{Physics-aware World Simulator for Robotic Learning}
Physical structure can be introduced through the control interface, the predicted state, or the training objective.
Visual robot trajectories and kinematic representations expose the spatial realization of actions~\cite{alzayer2026masked,kim2026robotfactored,xu2026kinema4d}; ContactFlow instead conditions manipulation prediction on three-dimensional contact-point trajectories shared across embodiments~\cite{azirar2026contactflow}.
Geometric motion representations also connect observation to dynamics: Point4D reconstructs long-range point trajectories from video~\cite{jeon2026point4d}, whereas PointWorld predicts action-conditioned scene changes as three-dimensional point flows~\cite{huang2026pointworld}.
These roles are distinct from supplying a future scene trajectory as a video-generation control.

FlowDreamer predicts scene flow to condition future RGB generation~\cite{guo2025flowdreamer}, while trajectory-based guidance aligns appearance features with specified motion~\cite{chu2025wanmove}.
Other approaches emphasize training signals, including joint prediction of visual and foundation-model states~\cite{tan2026dreamworld}, trajectory and relational alignment~\cite{zhang2026physisforcing}, and action-sensitive loss allocation~\cite{fang2026caer}.
Closely related, PHANTOM integrates physical-state inference into generation by jointly predicting video and V-JEPA2 latent physical representations with two flow-matching branches coupled through bidirectional cross-attention~\cite{shen2026phantom}.
JEPA Guided Diffusion instead decouples the two: a frozen V-JEPA encoder extracts predictive representations from the observed video, and a lightweight alignment module projects them into the conditioning space of a frozen Cosmos diffusion model for traffic forecasting~\cite{nguyen2026jepa}.
\method{} also conditions generation on V-JEPA features, but predicts them from the commanded robot actions and applies them selectively in space.
It uses optical flow as auxiliary training supervision and injects internally predicted V-JEPA feature changes through a learned motion gate and an environment mask.
This connects region-balanced motion supervision with spatially modulated predictive conditioning during RGB denoising.
This formulation targets faithful visual action consequences without equating optical flow with contact dynamics or a physically complete scene state.

\section{Proposed Method}
\label{sec:method}

\begin{figure*}[t]
 \centering
 \includegraphics[width=\textwidth]{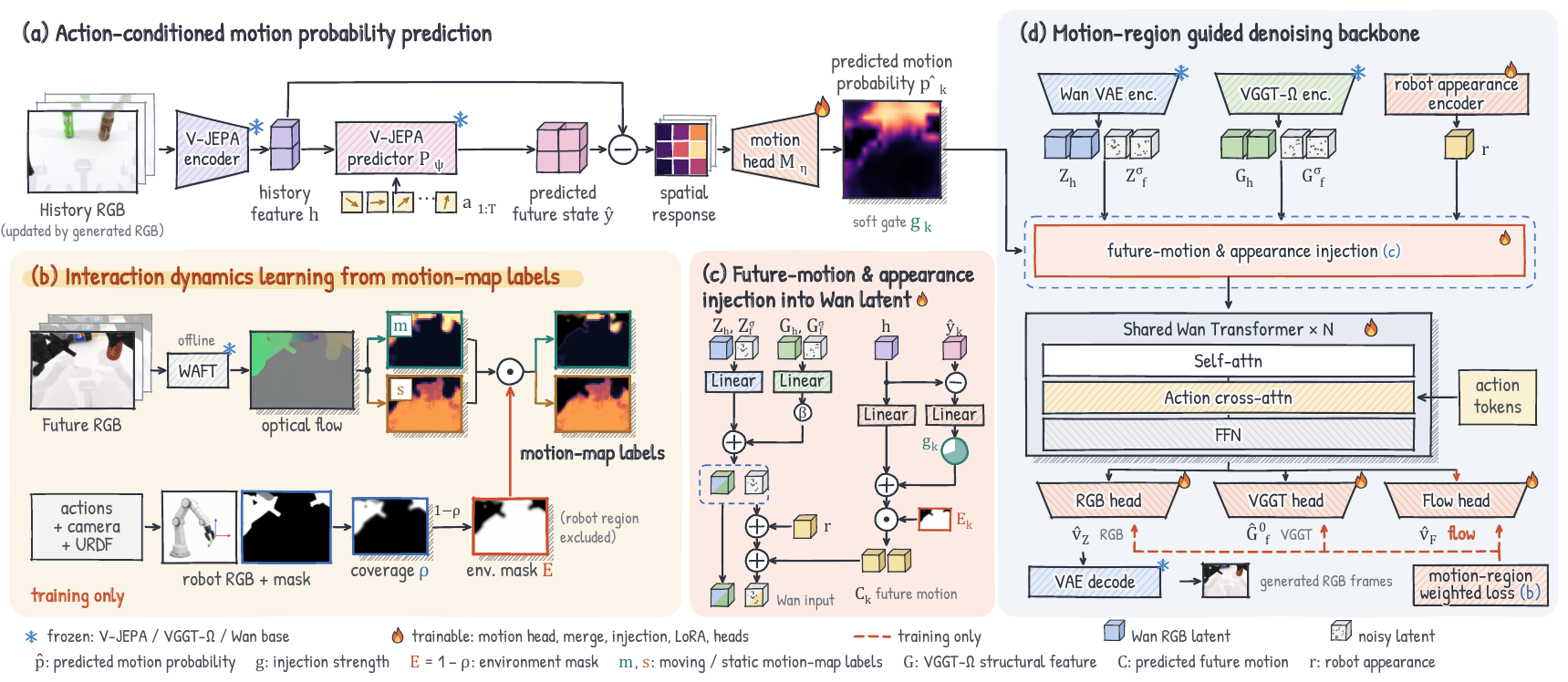}
 \caption{\textbf{An overview of MoReWorld.} (a) An action-conditioned V-JEPA predictor estimates the future state from history and time-aligned actions; a motion head converts its spatial response into a motion probability $\hat p_k$ and a soft gate $g_k$. (b) During training, WAFT optical flow yields moving/static labels $m,s$, and URDF rendering yields the environment mask $E=1-\rho$ that excludes the robot region; their product gives the motion-map labels. (c) The gated future motion $C_k$ and the robot appearance $r$ are injected into the Wan input. (d) A shared Wan denoising transformer jointly processes RGB and VGGT latents and predicts the RGB velocity, clean future VGGT features, and an auxiliary flow velocity, supervised with the motion-region weighted loss.}
 \label{fig:pipeline}
\end{figure*}

\paragraph{Problem setup.}
\method{} generates the $T$ frames that follow a history of $L$ RGB frames, given the robot state and a commanded action sequence $a_{1:T}$. During training, the history is taken from the recorded video. At inference, only a single initial image is observed: the first history is formed by repeating this image, and every later history consists of the model's own generated frames, so the video is rolled out autoregressively. We build on a latent video generator trained with flow matching~\cite{wan22}: a clean latent $x_0$ is perturbed to $x^\sigma=(1-\sigma)x_0+\sigma\epsilon$ with $\epsilon\sim\mathcal N(0,I)$, and the network regresses the velocity $v=\epsilon-x_0$. The video VAE compresses every four frames into one latent \emph{cell}, so the $T$ future frames form $K=T/4$ cells indexed by $k$.

\paragraph{Overview.}
A dense denoising loss is dominated by the static background and by the robot, whose motion is already specified by the action. The object response, which determines action-consequence fidelity, occupies few pixels and is therefore weakly supervised and weakly conditioned. \method{} addresses this with one spatial concept, the \emph{motion region}: the part of the environment, excluding the robot, that moves in response to the action. As shown in Fig.~\ref{fig:pipeline}, the motion region is (a) predicted as a \emph{motion probability} from history and actions before generation, (b) annotated by \emph{motion-map labels} from training videos, which supervise this prediction and re-weight the denoising losses, and (c) used to gate where predicted future features are injected into (d) a shared denoising backbone.

\subsection{Action-Conditioned Motion Probability Prediction}
\label{sec:action_motion_prediction}

Deciding \emph{where} the scene will respond requires anticipating the outcome of the action, but not rendering it. We therefore predict the response in the feature space of a frozen V-JEPA encoder~\cite{assran2025vjepa2}, where future prediction is deterministic and inexpensive (Fig.~\ref{fig:pipeline}(a)). The encoder maps each history frame to patch features $h_{1:L}$, and we write $h=h_L$ for the latest one. An action-conditioned predictor $P_\psi$ receives the history features together with per-frame action and robot-state tokens $q$, and predicts the features of all future frames in a single pass:
\begin{equation}
 \hat y_{1:T}=P_\psi\!\left(h_{1:L},\,a_{1:T},\,q\right).
 \label{eq:future_feature_prediction}
\end{equation}
$P_\psi$ is pretrained by regressing the encoder features of the true future frames and is kept frozen afterwards.

For cell $k$, covering frames $\mathcal T_k=\{4k-3,\dots,4k\}$, we summarize the prediction by its mean feature $\bar y_k$ and by the \emph{spatial response} $d_k$, the per-patch magnitude of the predicted change relative to the latest observation:
\begin{equation}
 \bar y_k=\frac14\sum_{t\in\mathcal T_k}\hat y_t,\qquad
 d_k=\frac14\sum_{t\in\mathcal T_k}\left|\hat y_t-h\right|.
 \label{eq:spatial_response}
\end{equation}
A lightweight motion head $M_\eta$ converts them into a motion probability map and a soft gate,
\begin{equation}
\begin{aligned}
 \hat p_k&=\operatorname{sigmoid}\!\big(M_\eta([h,\bar y_k,d_k])\big),\\
 g_k&=\alpha+(1-\alpha)\,\hat p_k .
\end{aligned}
 \label{eq:motion_probability}
\end{equation}
The floor $\alpha$ keeps a weak conditioning pathway in regions predicted to be static, so that an under-estimated motion region does not remove the future evidence altogether, $\alpha$ is set to 0.1. Both $\hat p_k$ and $g_k$ depend only on history and actions, and are thus available at inference.

\subsection{Interaction Dynamics Learning from Motion-Map Labels}
\label{sec:region_dynamics_learning}

Training videos reveal where the environment actually responded. We convert this evidence into motion-map labels (Fig.~\ref{fig:pipeline}(b)) that supervise $\hat p_k$ and re-weight the denoising losses.

\paragraph{Motion-region labels and loss.}
We compute optical flow between consecutive future frames offline with WAFT~\cite{wang2025waft} and discard pixels whose flow is unreliable. The remaining pixels are split by flow magnitude into moving and static ones, and pooling these indicators to the latent grid of each cell gives soft labels \(m_k, s_k \in [0,1]\). However, optical flow alone would mark the robot as the dominant moving region. Since robot motion is already determined by the action, we render the commanded trajectory with the robot URDF and the camera parameters to obtain a robot RGB image and mask for every frame. Pooling the mask within each cell gives a soft coverage \(\rho_k \in [0,1]\), and the environment mask is defined as \(E_k = 1 - \rho_k\). Unlike the flow-derived labels, \(E_k\) depends only on the commanded actions and is also computed at inference time.

Given the environment mask, the final motion-map labels are obtained by restricting the flow-derived labels to the environment:
\begin{equation}
 \tilde m_k=E_k\odot m_k,\qquad \tilde s_k=E_k\odot s_k .
 \label{eq:environment_supports}
\end{equation}
For a future prediction $\hat x$ and its target $x$, let $i$ index the spatial locations in the latent grid. We define the region loss as a weighted sum of two normalized squared errors:
\begin{equation}
\begin{aligned}
 \mathcal L_{\mathrm{reg}}(\hat x,x)
 ={}&(1-\gamma)\,\frac{\sum_i \tilde m_i\,\|\hat x_i-x_i\|_2^2}{\sum_i \tilde m_i}\\
 &+\gamma\,\frac{\sum_i \tilde s_i\,\|\hat x_i-x_i\|_2^2}{\sum_i \tilde s_i}.
\end{aligned}
 \label{eq:motion_region_objective}
\end{equation}
Because each term is normalized by the area of its own region, a small moving region is not outweighed by the large static background. A small $\gamma$ lets the moving environment dominate, while the static term still penalizes spurious object motion. Finally, $\mathcal L_{\mathrm{reg}}$ serves as an auxiliary term that is added to, rather than substituted for, the standard dense loss (Sec.~\ref{sec:training_objective}). This dual supervision preserves the fidelity of the robot's appearance and the global scene while strongly emphasizing the interaction regions.

\subsection{Future-Motion and Appearance Injection}
\label{sec:motion_gated_conditioning}

Fig.~\ref{fig:pipeline}(c) shows how the predicted motion enters the generator. For cell $k$, we use the predicted feature of its last frame, $\hat y_k\equiv\hat y_{4k}$, and form the \emph{future-motion} condition
\begin{equation}
 C_k=E_k\odot\Big[W_c\,\mathcal U(h)+g_k\odot W_d\,\mathcal U(\hat y_k-h)\Big],
 \label{eq:gated_dynamic_bridge}
\end{equation}
where $\mathcal U$ resamples V-JEPA patch features to the Wan token grid and $W_c,W_d$ are linear projections. The first term provides the current scene context; the second injects the predicted change, with strength set by the gate $g_k$. Hence predicted future features act mainly where the environment is expected to move, and inaccurate predictions in static regions have little influence on the generated frames. The mask $E_k$ removes the condition over the robot, whose appearance is supplied separately: a convolutional encoder maps the rendered robot RGB and mask of each cell to an appearance feature $r$. Keeping $r$ and $C_k$ on disjoint regions prevents the commanded robot motion from being confused with the predicted scene response.

Let $Z_h$ and $Z_f^\sigma$ denote the clean history and noisy future RGB latents from the Wan VAE, and $G_h$, $G_f^\sigma$ the corresponding structural features from a frozen VGGT-$\Omega$ encoder, which we incorporate following DreamWorld~\cite{tan2026dreamworld}. The input tokens of the denoiser are
\begin{equation}
 X^\sigma=\mathcal P_Z\big([Z_h,Z_f^\sigma]\big)
 +\beta\,\mathcal P_G\big([G_h,G_f^\sigma]\big)+r+[\,0,\,C\,],
 \label{eq:wan_input}
\end{equation}
where $\mathcal P_Z$ is the original Wan patch embedding, $\mathcal P_G$ is a linear embedding of the structural features scaled by $\beta$, and $[\,0,C\,]$ places $C_{1:K}$ on the future cells only.

\subsection{Motion-Region-Guided Denoising Backbone}
\label{sec:denoising_backbone}

For the backbone (Fig.~\ref{fig:pipeline}(d)), we follow the architecture of DreamWorld~\cite{tan2026dreamworld}, which integrates world-prior knowledge into a video generator by denoising RGB latents jointly with foundation-model features. A shared Wan transformer $T_\theta$ processes $X^\sigma$, with the text condition replaced by action tokens that enter every block through cross-attention. Three heads decode the future tokens:
\begin{equation}
 \big(\hat v_Z,\;\hat G_f^0,\;\hat v_F\big)=H\big(T_\theta(X^\sigma,\sigma,a_{1:T})\big).
 \label{eq:multihead_outputs}
\end{equation}
The RGB head predicts the velocity $\hat v_Z$ of the RGB latent, and the VGGT head predicts the clean future structural feature $\hat G_f^0$. What we add to this backbone is the motion-region guidance: the injected condition of Eq.~\ref{eq:gated_dynamic_bridge}, the region loss on every output, and a flow head. The flow head predicts the velocity $\hat v_F$ of a flow latent, obtained by encoding the WAFT flow with the Wan VAE. The noisy flow latent is seen only by this head and is never fed to the transformer, so flow acts purely as auxiliary supervision that makes the shared tokens predictive of local motion; the head is discarded at inference.

\subsection{Training Objective and Inference}
\label{sec:training_objective}

\paragraph{Objective.}
Let $\mathcal L_{\mathrm{mse}}(\hat x,x)$ denote the standard mean squared error over all locations, and let $v_Z$, $v_F$, and $G_f$ be the targets of the three heads. Each head combines the dense loss with the region loss of Eq.~\ref{eq:motion_region_objective}:
\begin{equation}
\begin{aligned}
 \mathcal L_{\mathrm{rgb}}&=\mathcal L_{\mathrm{mse}}(\hat v_Z,v_Z)+\lambda_r\,\mathcal L_{\mathrm{reg}}(\hat v_Z,v_Z),\\
 \mathcal L_{\mathrm{geo}}&=\mathcal L_{\mathrm{mse}}(\hat G_f^0,G_f)+\lambda_r\,\mathcal L_{\mathrm{reg}}(\hat G_f^0,G_f),\\
 \mathcal L_{\mathrm{flow}}&=\mathcal L_{\mathrm{reg}}(\hat v_F,v_F).
\end{aligned}
 \label{eq:head_losses}
\end{equation}
The motion probability is supervised by a binary cross-entropy against the motion-map labels, with the moving and static regions balanced in the same way:
\begin{equation}
 \mathcal L_{p}=-\frac12\left[
 \frac{\sum_i \tilde m_i\log \hat p_i}{\sum_i \tilde m_i}
 +\frac{\sum_i \tilde s_i\log (1-\hat p_i)}{\sum_i \tilde s_i}\right].
 \label{eq:motion_probability_loss}
\end{equation}
The full objective is
\begin{equation}
 \mathcal L=w(\sigma)\big(\mathcal L_{\mathrm{rgb}}+\lambda_F\mathcal L_{\mathrm{flow}}\big)
 +\lambda_G\mathcal L_{\mathrm{geo}}+\lambda_p\mathcal L_{p},
 \label{eq:full_objective}
\end{equation}
where $w(\sigma)$ is the flow-matching weight of the base model. The motion region thus plays three roles through a single spatial definition: its labels supervise where motion is predicted ($\mathcal L_p$), they re-weight what the generator is penalized for ($\mathcal L_{\mathrm{reg}}$), and its predicted probability, through the gate $g_k$, determines where future features are injected (Eq.~\ref{eq:gated_dynamic_bridge}).

\paragraph{Inference.}
Inference uses only causal inputs and requires no optical flow, motion-map labels, or future observations. For each segment, a single forward pass of \(P_\psi\) and \(M_\eta\) on the current history gives the predicted features \(\hat{y}\) and the gate \(g\), and rendering the commanded trajectory gives the appearance condition \(r\) and the environment mask \(E\); these conditions are fixed across the denoising steps. The RGB and structural latents are then jointly denoised, and the decoded frames form the history of the next segment.

\section{Experiments}
\label{sec:experiments}

\begin{table*}[t]
 \centering
 \caption{WorldSimProbe results on RoboTwin across five simulator-faithfulness capabilities. Scores are on a 0--100 scale, with higher values indicating greater faithfulness. Overall is the unweighted mean of the five suite scores. Bold and underlined values denote the best and second-best results, respectively.}
 \label{tab:worldsimprobe_comparison}
 \small
 \setlength{\tabcolsep}{8pt}
 \begin{tabular}{lrrrrrr}
  \toprule
  Method & T1 Local & T2 Global & T3 Source & T4 Ground. & T5 Dynamics & Overall \\
  \midrule

  Ctrl-World~\cite{guo2025ctrlworld} & \underline{52.40} & 51.10 & 48.80 & 62.00 & 25.48 & 47.96 \\
  DreamDojo~\cite{gao2026dreamdojo} & 35.20 & 45.70 & 39.20 & 49.00 & 26.90 & 39.20 \\

  Cosmos-3-Nano~\cite{agarwal2026cosmos} & 35.20 & 46.20 & 35.50 & 42.00 & 27.30 & 37.24 \\
  LingBot-VA~\cite{li2026causal} & 48.00 & \underline{64.40} & \textbf{62.00} & 56.00 & \underline{44.23} & 54.93 \\
  MVA~\cite{alzayer2026masked} & 50.12 & 63.70 & 51.30 & \textbf{70.00} & 41.50 & \underline{55.32} \\

  \method{} (Ours) & \textbf{59.19} & \textbf{64.79} & \underline{52.70} & \underline{66.00} & \textbf{63.38} & \textbf{61.21} \\

  \bottomrule
 \end{tabular}
\end{table*}

\begin{figure*}[t]
 \centering
 \includegraphics[width=\textwidth]{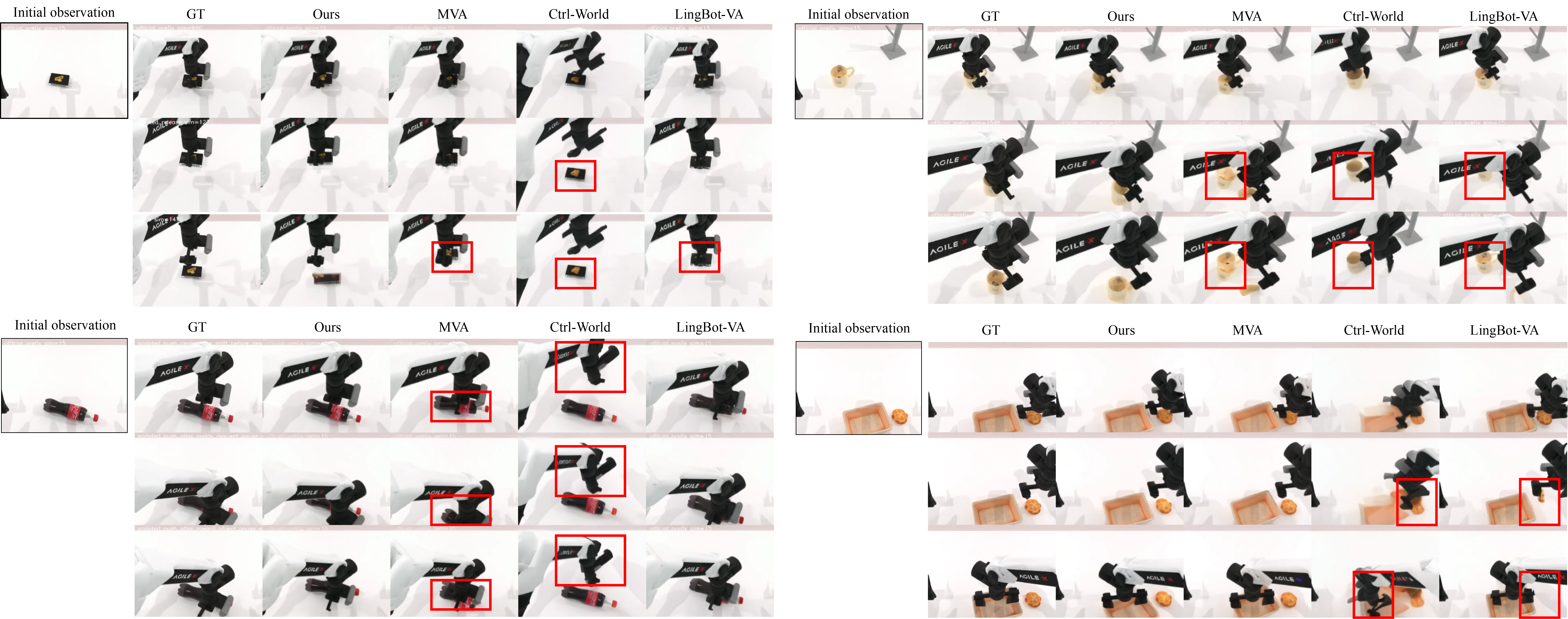}
 \caption{\textbf{Qualitative comparison on WorldSimProbe.} Each panel shows the initial frame (left) and three later time steps (rows) of one benchmark case. Columns show, from left to right, the ground truth, \method{} (Ours), MVA, Ctrl-World, and LingBot-VA. All models receive the same initial frame and action trajectory. Red boxes mark incorrect object responses or robot motion.}
 \label{fig:qualitative}
\end{figure*}

\subsection{Implementation Details}
\label{sec:experimental_setup}

\paragraph{Dataset.}
We build the training set on RoboTwin~2.0~\cite{chen2025robotwin}: 13,310 episodes, about 48.85 hours at 50\,Hz, with 14-dimensional dual-arm joint actions.
Of these, 5,000 are clean demonstrations and the rest are collected by executing $\pi_{0.5}$~\cite{black2025pi05}; episodes are not filtered by outcome, so both successful and failed executions are included.

\paragraph{Model and training.}
\method{} builds on Wan2.2-TI2V-5B~\cite{wan22}, with a frozen V-JEPA~2.1 ViT-L encoder~\cite{assran2025vjepa2}, a frozen VGGT-$\Omega$, and WAFT~\cite{wang2025waft} flow labels.
Each window has 16 history and 32 future frames at 25\,Hz and $640\times480$ resolution.
We train for 30,000 updates with AdamW and a batch size of four on eight RTX~4090 GPUs, using $\gamma=0.05$, $\lambda_r=0.25$, and $\lambda_F=\lambda_G=\lambda_p=0.1$.

\subsection{Quantitative results}
\label{sec:model_comparison}

\paragraph{Benchmark and metrics.}
We evaluate on the RoboTwin subset of WorldSimProbe~\cite{co2026worldsimprobe}, which scores simulator faithfulness with five suites on a 0--100 scale.
\textbf{Local}, \textbf{Global}, and \textbf{Source} test whether the commanded actions are realized as robot motion under action perturbations, trajectories replayed from other tasks, and different control sources, respectively.
\textbf{Ground.} is the fraction of no-contact rollouts in which the object stays stationary, and \textbf{Dynamics} measures how often the generated interaction matches the intended primitive among eight classes.
Overall is their unweighted mean.
Ground.\ uses the official evaluator; for Dynamics, we keep the official prompt and motion gates and use GPT-6.1-sol as the judge.

\paragraph{Compared methods.}
We compare with five action-conditioned world models: Ctrl-World~\cite{guo2025ctrlworld}, DreamDojo~\cite{gao2026dreamdojo}, Cosmos-3-Nano~\cite{agarwal2026cosmos}, LingBot-VA~\cite{li2026causal,lingbotva_implementation}, and Masked Visual Actions (MVA)~\cite{alzayer2026masked}, which, like \method{}, is conditioned on robot renderings.
Ctrl-World and LingBot-VA are adapted from their official implementations with rank-32 LoRA and a 14-dimensional action interface.

\begin{table*}[t]
 \centering
 \caption{Component studies on the same 200 WorldSimProbe videos. Visual metrics are computed over both tasks; interaction scores are reported separately. $\Delta$T4 and $\Delta$T5 are relative to the MoReWorld full model, computed as ablation minus full-model score; negative values indicate a decrease. Removing V-JEPA excludes the entire predictive-conditioning branch. Removing the flow head also removes its training loss. The w/o motion-region variant removes motion-region effects from region weighting and motion-gated conditioning while retaining the flow head and VGGT. Bold denotes the best value per column.}
 \label{tab:component_ablation}
 \small
 \setlength{\tabcolsep}{9pt}
 \begin{tabular}{lrrrrrrrr}
  \toprule
  Method & LPIPS$\downarrow$ & $\mathrm{FVD}_{16}\downarrow$ & PSNR$\uparrow$ & SSIM$\uparrow$ & T4 Ground.$\uparrow$ & $\Delta$ T4 & T5 Dynamics$\uparrow$ & $\Delta$ T5 \\
  \midrule
  w/o V-JEPA & 0.1387 & 109.38 & 20.04 & 0.8608 & 48.00 & \textcolor{red}{\textbf{$-18.00$}} & 59.21 & $-4.17$ \\
  w/o flow head & 0.1336 & 118.10 & 20.12 & 0.8626 & 54.00 & $-12.00$ & 58.01 & $-5.37$ \\
  w/o VGGT & 0.1204 & 93.19 & 20.66 & 0.8714 & 62.00 & $-4.00$ & 53.85 & $-9.53$ \\
  w/o motion-region & \textbf{0.1160} & \textbf{90.99} & \textbf{20.80} & \textbf{0.8737} & 64.00 & $-2.00$ & 30.45 & \textcolor{red}{\textbf{$-32.93$}} \\
  \midrule
  \method{} (full) & 0.1289 & 99.90 & 20.27 & 0.8652 & \textbf{66.00} & 0.00 & \textbf{63.38} & 0.00 \\
  \bottomrule
 \end{tabular}
\end{table*}

\begin{figure}[t]
 \centering
 \includegraphics[width=\linewidth]{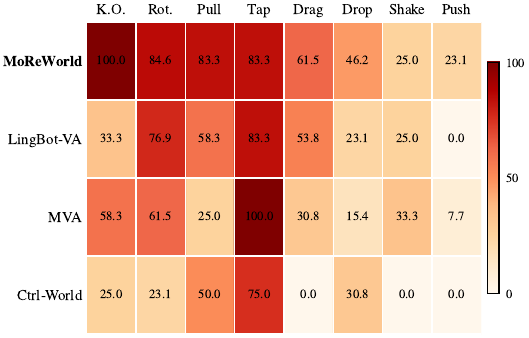}
 \caption{Dynamics score by interaction primitive (\%). K.O.\ denotes knock-over and Rot.\ rotate. Darker cells indicate higher scores.}
 \label{fig:task5_analyse}
\end{figure}

\paragraph{Results.}
Table~\ref{tab:worldsimprobe_comparison} reports the five suite scores.
\method{} obtains the highest Overall score, 61.21, which is 5.89 points above the second-best model, MVA; it ranks first on Local, Global, and Dynamics, and second on Source and Ground.
The largest gain is on Dynamics, where its score of 63.38 is 19.15 points above the strongest baseline, LingBot-VA.
Figure~\ref{fig:task5_analyse} shows that this gain is broad: \method{} is the best model on six of the eight primitives and reaches 100.0 on knock-over, the failure illustrated in Fig.~\ref{fig:motivation}(a), while push and shake remain difficult for all models.
The gain does not come from moving objects indiscriminately: on Ground., which rewards keeping untouched objects still, \method{} scores 66.00, 4 points below the best model, MVA, whose Dynamics score is 21.88 points lower.
For action realization, \method{} leads Local by 6.79 points and matches the best baselines on Global, where the top three models lie within 1.1 points.
On Source, it is second to LingBot-VA (52.70 vs.\ 62.00): motion-region guidance targets the environment response rather than the robot's own motion and does not close this gap.

\begin{figure*}[t]
 \centering
 \includegraphics[width=\textwidth]{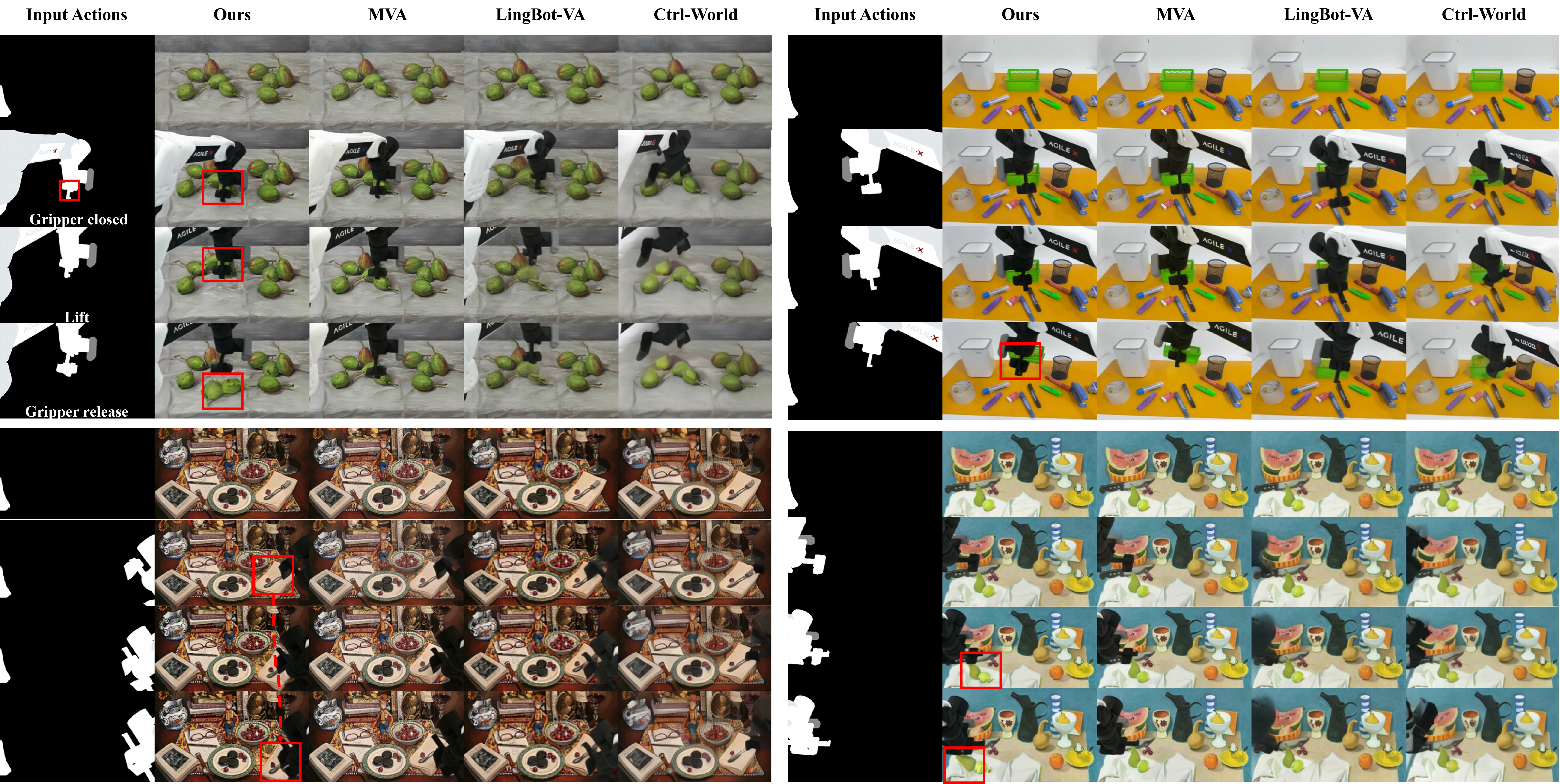}
 \caption{\textbf{Qualitative transfer to in-the-wild images.} The four panels show a pear painting (top left), a real tabletop photograph (top right), an illustrated still life (bottom left), and another painted still life (bottom right). Within each panel, columns show input actions as black-background URDF renderings, \method{} (Ours), MVA, LingBot-VA, and Ctrl-World, from left to right. The top row shows the supplied initial image; subsequent rows advance in time. Red boxes and the dashed arrow highlight selected object responses. Robot-specific adaptation uses simulated interactions, with no additional training on these external images.}
 \label{fig:in_the_wild}
\end{figure*}

\subsection{Qualitative results}
\label{sec:Qualitative results}

Figure~\ref{fig:qualitative} compares rollouts on four benchmark cases under the same initial frame and action trajectory.
The baselines fail in several ways, marked by red boxes: the object stays attached to the gripper after it is released (top left), is duplicated or deformed while being moved (top right), breaks apart under contact (bottom left), or disappears although it is never touched (bottom right); in some rollouts the arm also deviates from the commanded trajectory.
\method{} produces the appropriate response in all four cases: it leaves the released object on the table, moves a single upright mug with the gripper, keeps the pushed bottle intact, and leaves the untouched bread roll in place, although the released object is distorted in appearance.

\section{Analysis and Discussion}
\label{sec:analysis}

\subsection{Component Analysis}
\label{sec:component_studies}

Table~\ref{tab:component_ablation} removes one component at a time and evaluates each variant on the same 200 Ground.\ and Dynamics videos.

\paragraph{Motion-region guidance.}
Removing the motion region from the region-weighted losses and from the gated conditioning causes the largest change in the table: Dynamics falls from 63.38 to 30.45, while Ground.\ changes by only 2 points.
The same variant obtains the best LPIPS, $\mathrm{FVD}_{16}$, PSNR, and SSIM.
Pixel-level metrics are dominated by the static background and the robot, since moving pixels cover less than 2\% of the non-robot area.
A model can therefore reproduce the scene appearance while missing the object response, and the motion region is the component that directs learning toward that response.

\paragraph{Predictive conditioning, flow, and geometry.}
Removing the V-JEPA branch causes the largest drop on Ground., from 66.00 to 48.00, together with the worst LPIPS, PSNR, and SSIM.
This suggests that the predicted future features mainly tell the generator where the scene should not respond.
Removing the flow head lowers Ground.\ to 54.00 and Dynamics to 58.01 and gives the worst $\mathrm{FVD}_{16}$ (118.10), so flow supervision benefits both the interaction scores and temporal consistency.
Removing VGGT mainly affects Dynamics, which drops by 9.53 points, while the visual metrics improve slightly.
We also replace the gate $g=0.1+0.9p$ with $g=p$ at inference on the same checkpoint: Ground.\ drops to 62.00 and Dynamics to 60.18, so weak conditioning in regions predicted to be static remains useful.
The full model is the best on both interaction suites.

\subsection{Limitations}
\label{sec:limitations}

Table~\ref{tab:component_ablation} also shows the cost of motion-region guidance: the full model is worse than the w/o motion-region variant on all four visual metrics (e.g., LPIPS 0.1289 vs.\ 0.1160 and $\mathrm{FVD}_{16}$ 99.90 vs.\ 90.99).
In addition, \method{} tends to deform the objects it moves, as with the released object in Fig.~\ref{fig:qualitative} (top left).
We attribute both effects to the same cause.
The region loss and the gated injection tell the generator where and how strongly the scene should change, but neither constrains the shape of the object: the region loss is a per-location error, and the injected V-JEPA features lie on a $16\times16$ grid, which is too coarse to carry fine geometry and texture.
The generator is thus pushed to change the moving region without an object-level constraint, and such errors may accumulate over the autoregressive rollout.
Object-level constraints, such as rigidity or instance-consistent appearance, and higher-resolution predictive features are natural directions for reducing this deformation.

\subsection{In-the-Wild Generalization}
\label{sec:in_the_wild}

Figure~\ref{fig:in_the_wild} applies the models to a real tabletop photograph and three paintings or illustrations, using joint trajectories from simulated data and no training on these images.
In all four scenes, \method{} moves the highlighted object together with the gripper: it lifts a pear and leaves it at a new position after release (top left), displaces the green basket (top right), drags the napkin and fork along the dashed arrow (bottom left), and pushes the pear and apple aside (bottom right).
The baselines render the robot arm but mostly leave the objects in place or distort them; MVA also moves the basket in the photograph.
Since robot-specific adaptation uses only simulated interactions, these results suggest that the learned relation between robot motion and object response is not tied to the simulator's appearance.
The cases were selected after inspecting generated rollouts and have no ground-truth futures, so they illustrate possible transfer and do not measure how often it occurs.

\section{Conclusion}
\label{sec:conclusion}

In this work, we presented \method{}, a motion-region-guided framework that improves action-consequence fidelity in robotic video world models. We identify that existing methods often optimize global reconstruction quality while providing limited supervision for localized environmental responses. To address this issue, we introduced a unified Motion-Region Guidance paradigm that predicts where the environment will respond and uses this representation for motion prediction, region-aware supervision, and motion-gated future-feature conditioning. \method{} achieves the highest overall simulator-faithfulness score on the RoboTwin subset of WorldSimProbe and substantially improves Interaction Dynamics over existing baselines, demonstrating the effectiveness of spatially organized guidance for action-conditioned interaction modeling.
{\small
\bibliographystyle{ieeenat_fullname}
\bibliography{references}
}
\end{document}